\documentclass[]{spie}  

\usepackage{amsmath,amsfonts,amssymb}
\usepackage{graphicx}
\usepackage[colorlinks=true, allcolors=blue]{hyperref}
\usepackage{subfig}
\usepackage{tabularx}

\title{Learning to Guide Multiple Heterogeneous Actors from a Single Human Demonstration via Automatic Curriculum Learning in StarCraft II}

\author[a,b]{Nicholas Waytowich}
\author[a]{James Hare}
\author[a]{Vinicius G. Goecks}
\author[a]{Mark Mittrick}
\author[a]{John Richardson}
\author[a]{Anjon Basak}
\author[a]{Derrik E. Asher}
\affil[a]{Army Research Laboratory, Aberdeen Proving Ground, MD, USA}
\affil[b]{Columbia University, New York, NY, USA}

\authorinfo{Further author information: (Send correspondence to Nicholas Waytowich)\\  Nicholas Waytowich: nicholas.r.waytowich.civ@army.mil}

\begin{document} 
\maketitle

\begin{abstract}



Traditionally, learning from human demonstrations via direct behavior cloning can lead to high-performance policies given that the algorithm has access to large amounts of high-quality data covering the most likely scenarios to be encountered when the agent is operating.
However, in real-world scenarios, expert data is limited and it is desired to train an agent that learns a behavior policy general enough to handle situations that were not demonstrated by the human expert.
Another alternative is to learn these policies with no supervision via deep reinforcement learning, however, these algorithms require a large amount of computing time to perform well on complex tasks with high-dimensional state and action spaces, such as those found in \textit{StarCraft II}.
Automatic curriculum learning is a recent mechanism comprised of techniques designed to speed up deep reinforcement learning by adjusting the difficulty of the current task to be solved according to the agent's current capabilities.
Designing a proper curriculum, however, can be challenging for sufficiently complex tasks, and thus we leverage human demonstrations as a way to guide agent exploration during training.
In this work, we aim to train deep reinforcement learning agents that can command multiple heterogeneous actors where starting positions and overall difficulty of the task are controlled by an automatically-generated curriculum from a single human demonstration.  
Our results show that an agent trained via automated curriculum learning can outperform state-of-the-art deep reinforcement learning baselines and match the performance of the human expert in a simulated command and control task in \textit{StarCraft II} modeled over a real military scenario.
\end{abstract}


\section{INTRODUCTION}
Deep reinforcement learning (RL) has yielded many recent successes in solving a variety of complex problems ranging from video games, self-driving vehicles, robotic manipulators as well as various other simulated and real scenarios \cite{Mnih2015a, andrychowicz2018learning, lillicrap2015continuous}. 
RL, however, is a trial-and-error based method that is inherently sample inefficient and often requires large computational and data resources to learn reliable policies. The trial-and-error nature of RL results in random behaviors at the beginning of training, which makes RL poorly suited for certain tasks that require more guarantees on performance, such as physically embodied tasks.  Additionally, traditional RL approaches are typically very sample inefficient and have slow convergence rates. This is even more problematic in tasks with either large state and action spaces or environments with relatively long time horizons, which can require tens of millions of training samples \cite{Bellemare2013, mnih2013playing}.

One approach for overcoming these limitations is learning from example behavior or demonstrations of the desired task from expert policies. These expert policies can come from either some other autonomous agent or directly from a human. 
Traditionally, learning from human demonstrations via direct behavior cloning can lead to high-performance policies with significantly less computation time given that the algorithm has access to high-quality demonstration data covering the most likely scenarios to be encountered when the agent is operating. 
However, in real-world scenarios, this is often not the case as expert data is limited and thus most imitation learning techniques that learn from demonstrations, often fail when the agent encounters new scenarios that were never seen before in the training data due to the distributional drift problem \cite{Levine2019}. Since we desire that our artificial intelligence (AI) agents learn policies robust enough to handle situations that were not demonstrated by the human expert, relying solely on demonstrations that cover the span of possible scenarios is simply not feasible. 

One possible solution to improve learning efficiency for RL agents is to break-down complex tasks into a series of easier ones to solve problems that form a curriculum. 
Curriculum learning is a relatively new technique designed to speed up reinforcement learning by adjusting the difficulty of the task according to the agent's current capabilities. 
Curriculum learning in AI draws inspiration from the curriculum learning techniques that humans use on a daily basis to teach complex tasks and concepts. In the field of reinforcement learning, curriculum learning has shown great promise in improving the speed and effectiveness of RL agents \cite{salimans2018learningmontezuma,rudin2021learningtowalk,miki2022wyldanimal}. The main challenge in curriculum learning is how to design a proper curriculum for a complex problem. 
Our solution is to leverage a technique called \emph{Automatic Curriculum Learning} (ACL) from human demonstrations in which we attempt to extract a reasonable curriculum automatically from expert human demonstrations of the task \cite{salimans2018learningmontezuma}. Using human demonstrations as an implicit curriculum, we can incrementally allow the agent to learn from different portions of the demonstration, starting from a simpler task (i.e. near the end of the demonstration) and eventually building the agent's ability to perform more and more complex tasks. 

\begin{figure}
    \centering
    \includegraphics[width=0.75\textwidth]{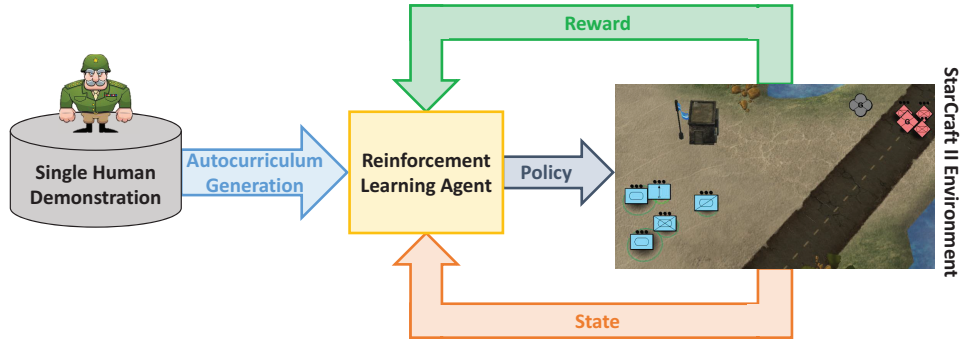}
    \caption{High-level overview of the proposed method. A single human demonstration is used to automatically generate a curriculum for the reinforcement learning agent to solve complex tasks within the challenging and high-dimensional state-space of \emph{StarCraft II}.}
    \label{fig:overall_concept}
\end{figure}

In this paper, we improve upon ACL, previously only  demonstrated in simple Atari tasks \cite{salimans2018learningmontezuma}, in which we use a single human demonstration to define an automatic curriculum to guide the exploration of an actor-critic RL agent to solve complex tasks within the challenging and high-dimensional state-space of \emph{StarCraft II}, as illustrated in Fig.~\ref{fig:overall_concept}.
We train deep reinforcement learning agents that can command multiple heterogeneous actors using an automatic curriculum learning technique where starting positions and overall difficulty of the task are controlled using a single human demonstration.  
Our results show that an agent trained via ACL outperform state-of-the-art deep reinforcement learning baselines and match the performance of the human expert in a simulated command and control task in \textit{StarCraft II} modeled over a real military scenario.

\section{METHODS}

\subsection{StarCraft II Environment}

\textit{StarCraft II} has a number of difficult challenges for AI algorithms that make it a suitable simulation environment for experimentation and development of deep reinforcement learning agents. For example, the game has notoriously complex state and action spaces, can last tens of thousands of time-steps, can have thousands of actions selected in real time, and can capture uncertainty due to the partial observability or ``fog-of-war''. Further, the game has multiple heterogeneous assets, that make learning infeasible for most traditional off-the-shelf RL techniques.

In this paper, we utilize a custom made \textit{StarCraft II} map that we developed in a previous paper \cite{dsifirstyear,goecks2022ongames} called ``TigerClaw". TigerClaw is a tactical command and control scenario where Blue Force units must engage and eliminate the Red Force units that currently occupy a neutral city. 
To facilitate training our RL agents, we developed a custom gym wrapper built around Deepmind's  \textit{StarCraft II Learning Environment} (SC2LE) \cite{vinyals2017starcraft}. To implement off-the-shelf deep reinforcement learning algorithms we utilize \textit{RLlib} \cite{liang2018rllib}, a library that provides scalable software primitives and high-quality reinforcement learning algorithms on a high performance computing system.
\textit{DeepMind}’s SC2LE framework \cite{vinyals2017starcraft} exposes \textit{Blizzard Entertainment}'s \textit{StarCraft II} Machine Learning API as a reinforcement learning environment. This tool provides access to \textit{StarCraft II}, its associated map editor, and an interface for RL agents to interact with \textit{StarCraft II}, getting observations and sending actions.

\begin{figure}
    \centering
    \subfloat[\label{fig:TigerClaw_a}]{\includegraphics[width=0.255\textwidth]{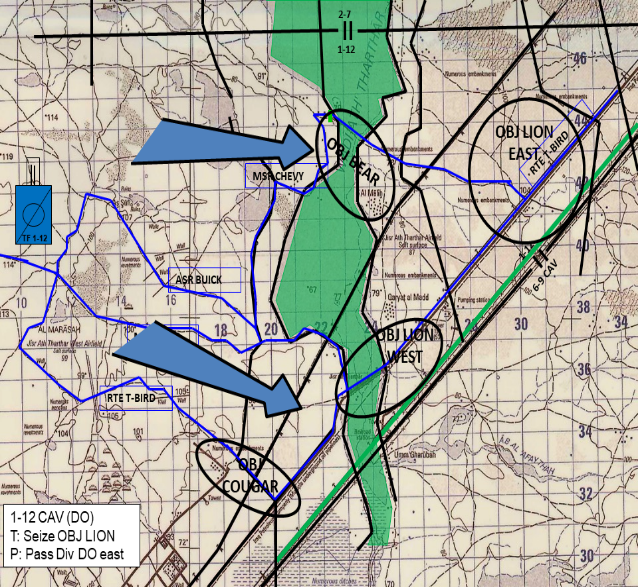}}\hfill
    \subfloat[\label{fig:TigerClaw_b}]{\includegraphics[width=0.405\textwidth]{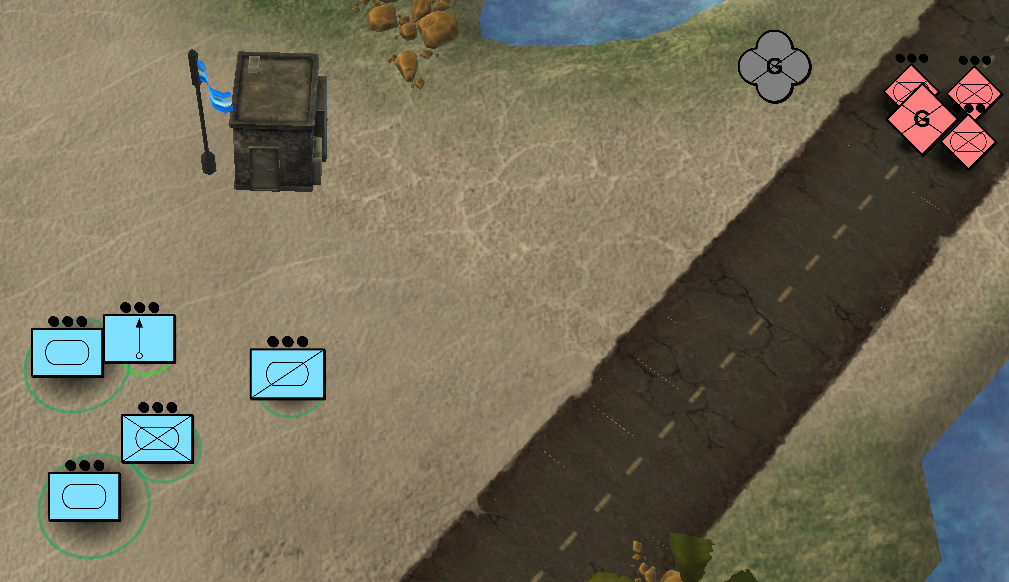}}\hfill
    \subfloat[\label{fig:TigerClaw_c}]{\includegraphics[width=0.301\textwidth]{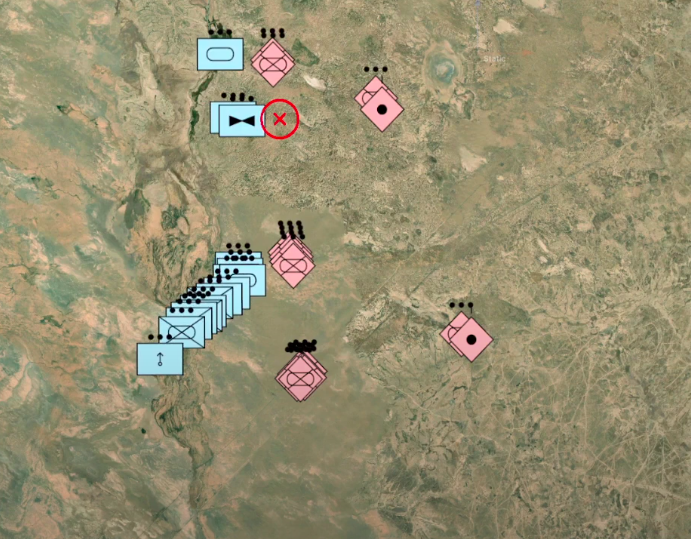}}
    \caption{Illustration of the \textit{TigerClaw} scenario map (Fig.~\ref{fig:TigerClaw_a}), how it was represented in \textit{StarCraft II} using MILSTD2525 symbols for the units (Fig.~\ref{fig:TigerClaw_b}), and the human-computer interface used to visualize the trajectories generated by both humans and learning agents in the Accelerated User Reasoning for Operations, Research, and Analysis (AURORA) \cite{dennison2020accelerated} virtual reality platform (Fig.~\ref{fig:TigerClaw_c}).}
    \label{fig:TigerClaw}
\end{figure}


\subsubsection{Task: The TigerClaw Scenario} \label{sec:tigerclaw}
In \textit{TigerClaw}, the Blue Force's goal is to cross a dry riverbed (wadi) terrain, neutralize the Red Force, and control certain geographic locations. Fig.~\ref{fig:TigerClaw} shows the overall objective of the TigerClaw Scenario. These objectives are encoded in the game score for use by reinforcement learning agents as a baseline for comparison across different neural network architectures and reward driving attributes. 

\subsubsection{State and Action spaces} \label{sec:state_action_space}

The original \textit{StarCraft II} state-space consisted of approximately 20 images of size 64x64 (13 screen feature layers and 7 mini-map feature layers) which included categorical features ranging from unit type to geographical path traversal information \cite{Vinyals2017}.
In this tactical version of the \textit{StarCraft II} mini-game, we utilize a custom state-space consisting of two representations: (i) \emph{image representation} and (ii) \emph{vector representation}. The image representation consists of a single 256 x 256 image of the entire battlefield that includes all of the relevant information of the original feature maps as well as terrain information, as seen in Fig.~\ref{fig:TigerClaw_c}. The vector representation is a set of $287$ non-spatial features that include all of the information about the game and units, including unit type, position, health, player resources, and build queues.

The actions in \textit{StarCraft II} are compound actions in the form of functions that require arguments and specifications about what unit should take an action and where that action is intended to take place on the screen. For example, once a unit is selected, an action such as ``attack'' is represented as a function that would require the $x-y$ attack locations on the screen. The action space consists of the action identifier (i.e., which action to run), and two spatial actions ($x$ and $y$) that 
represent the pair of coordinates in the screen where the action should be executed. This results in a very large action space that is impractical to be represented in a flattened space. To reduce this complexity, we defined \emph{cardinal actions} that divide the map into a $C$x$C$ grid, where we set $C=3$. Thus, the possible actions for the $x$ and $y$ coordinates are represented as two vectors of length $C=3$ with real-valued entries between $0$ and $1$ constituting left, center, and right, and top, middle, and bottom, respectively.

\subsubsection{Game Score and Reward Implementation}

The reward function is an important component of reinforcement learning as it ultimately controls how the agent reacts to environmental changes by giving them positive or negative reward for each situation. We developed a custom reward function for the \textit{TigerClaw} scenario consisting of awarding $+10$ points for the Blue Force crossing the wadi (river) and $-10$ points for retreating back. In addition, we awarded $+10$ points for destroying a Red Force unit and $-10$ points if a Blue Force unit was destroyed. The overall goal of this reward function was to incentivize the elimination of the opposing force while preserving Blue Force units from being destroyed.

\subsection{Reinforcement Learning Agent}

Next, we developed and implemented a modern deep reinforcement learning agent into the \textit{StarCraft II} environment to achieve the goal of the TigerClaw scenario discussed in Section~\ref{sec:tigerclaw}. We designed the action-space of our agent by adapting the traditional control approach of \textit{StarCraft II}, where the player performs the following steps to control each unit: (1) Select a unit using their mouse pointer, (2) Select an action based on the unit's possible actions and observed state, and (3) Select the $(x,y)$ coordinate in which the unit will execute the selected action.

\subsubsection{Learning Algorithm}
To train our deep reinforcement learning agent, we used the Asynchronous Advantage Actor Critic (A3C) algorithm, which is a state-of-the-art on-policy RL algorithm shown to have success on numerous challenging environments \cite{mnih2016asynchronous}. The A3C is a distributed RL algorithm that allows for parallelized exploration and training across multiple actors simultaneously. The A3C algorithm is an extension of the Advantage Actor Critic (A2C) in which multiple agents explore parallel environments simultaneously to speed up exploration and learning. Just like the A2C, the A3C maintains a a policy $\pi(a_t|s_t;\theta)$ and an estimate of the value function $V(s_t;\theta_v)$. In A3C, multiple copies of the actor policy are distributed across multiple instances of the environment to speed up exploration. 

Our A3C model was trained with $35$ parallel actor-learners on separate threads for over $70$ million timesteps (around $112$ thousand simulated battles) against a built-in \textit{StarCraft II} bot operating on hand crafted rules.
Each trained model was tested on 100 rollouts of the agent on the \textit{TigerClaw} scenario. The curriculum learning models are compared against a traditional baseline A3C approach with details in the Experiments and Results section.

\subsubsection{Network Architecture} 

\begin{figure}
    \centering
    \subfloat[Overall Network Architecture \label{fig:architecture}]{\includegraphics[width=0.65\textwidth]{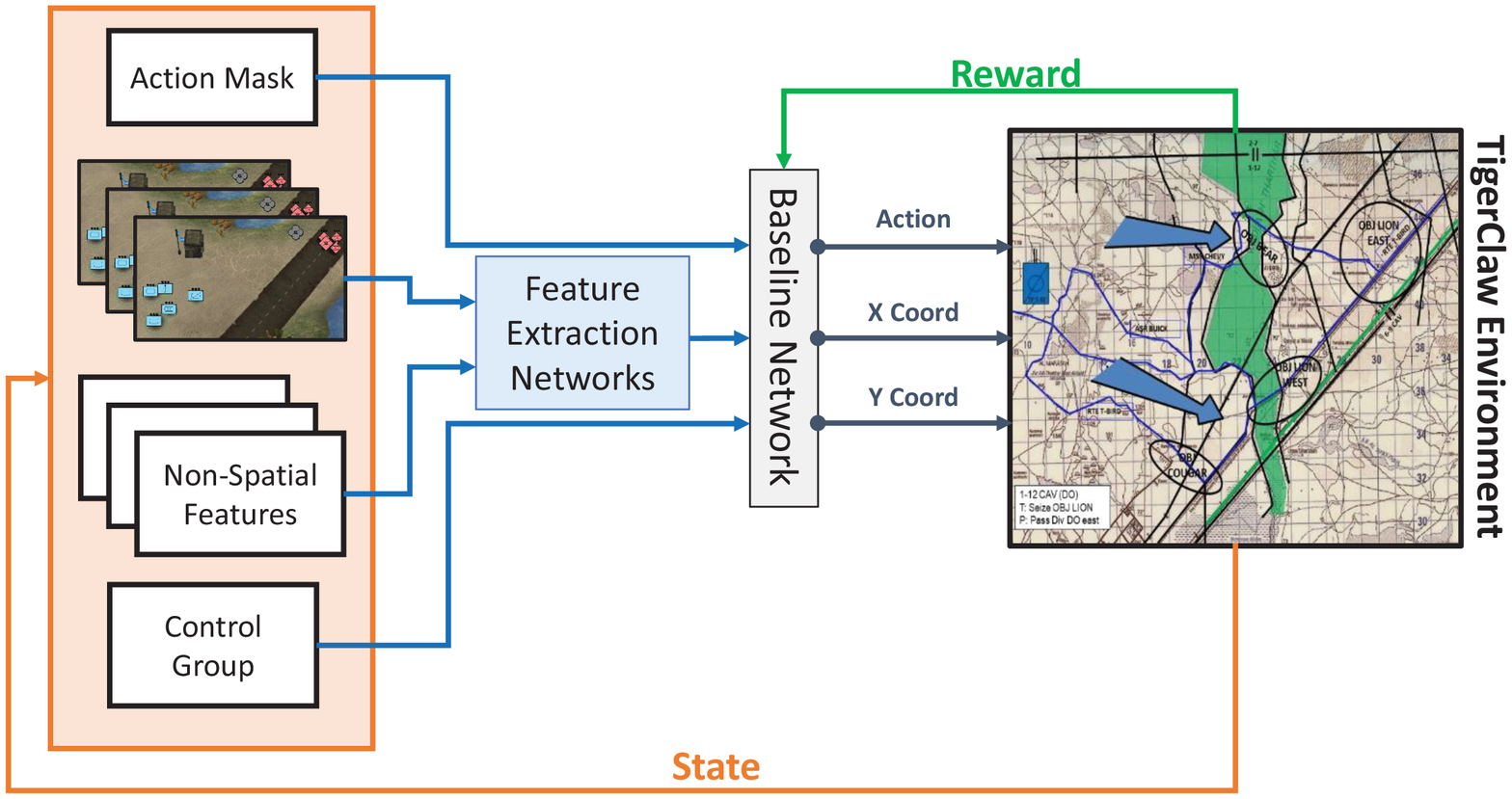}}\hfill
    \subfloat[``Feature Extraction Networks" Block \label{fig:features}] {\includegraphics[width=0.45\textwidth]{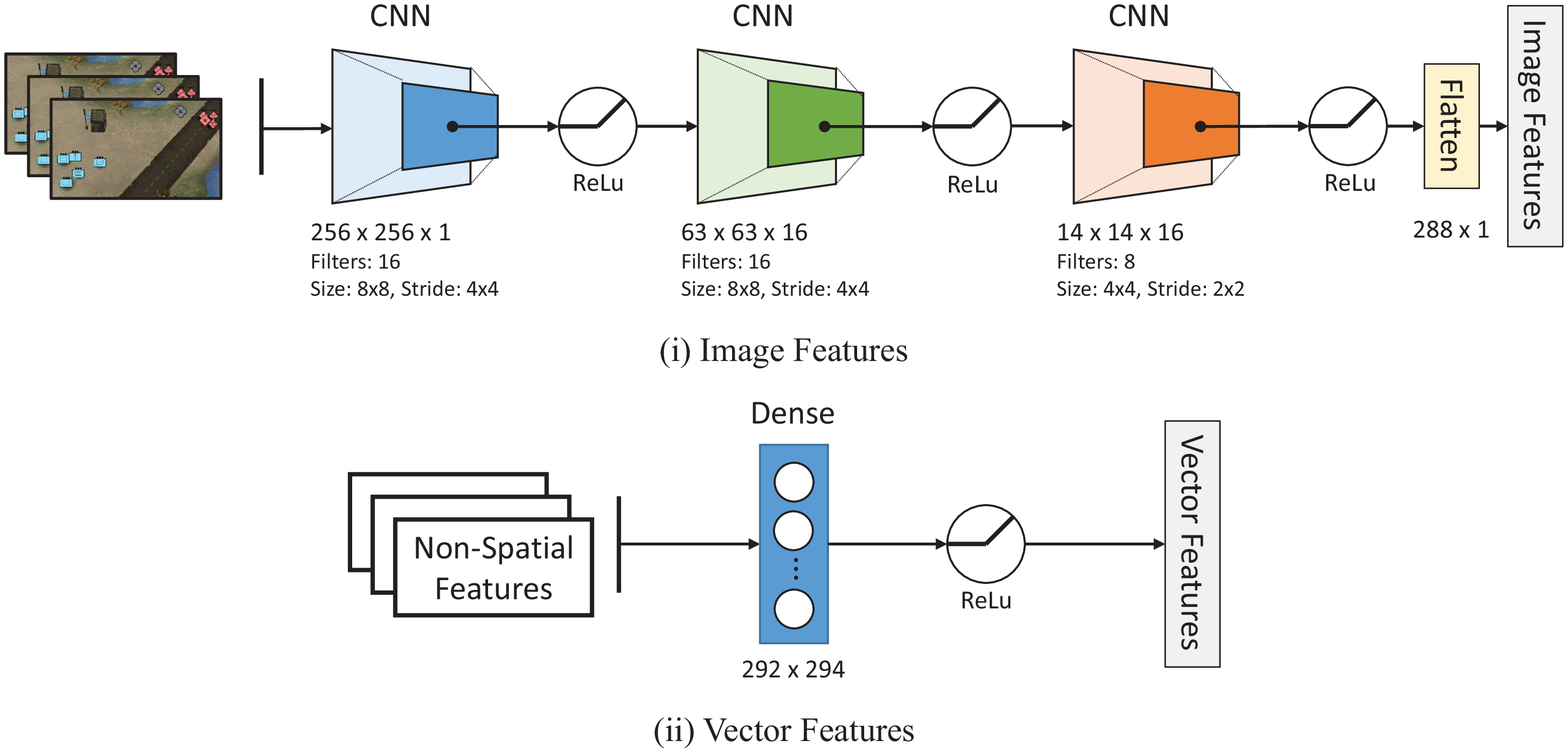}}\hfill
    \subfloat[``Baseline Network" Block\label{fig:baseline}]{\includegraphics[width=0.45\textwidth]{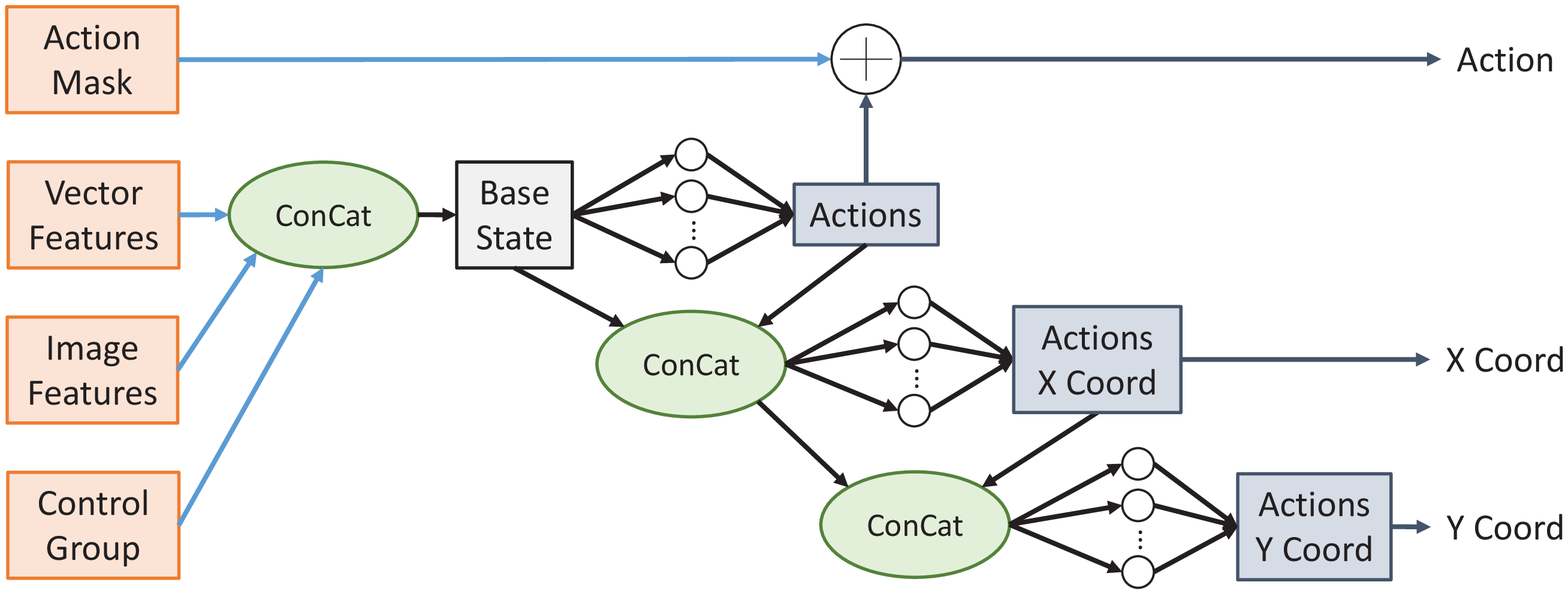}}
        \caption{Diagrams illustrating the overall network architecture of the reinforcement learning agent (Fig.~\ref{fig:architecture}), the architecture of the networks responsible to learn the main features of the environment (Fig.~\ref{fig:features}), and the network used to select all components of the actions (Fig.~\ref{fig:baseline}).}
        \label{fig:approach}
\end{figure}

As shown in Fig.~\ref{fig:architecture}, our deep reinforcement learning agent is represented by a multi-input and multi-output neural network adapted from Waytowich et al 2019 \cite{waytowich2019ICML} capable of handling the complex state-space and the multi-faceted action-space of \textit{StarCraft II}.


The state observations, as described in Section~\ref{sec:state_action_space}, are fed through a feature extraction network, shown in Fig.~\ref{fig:features}, to identify all of the relevant information available. Next, the state-space is fed into the Baseline network, shown in Fig.~\ref{fig:baseline}, which outputs the action and $(x,y)$ coordinates of the control group. The agent then executes the policy into the TigerClaw environment and receives a reward and a new state-space. The details of the state-space, feature extraction network, and the baseline network are described in the following subsections.
As shown in Fig.~\ref{fig:features}.i, the screen features were processed through three Convolutional Neural Network (CNN) layers with ReLu activation functions in order to extract visual feature representations of the global and local states of the map, respectively. 
The non-spatial features were processed through a fully-connected layer with a non-linear activation as shown in Fig.~\ref{fig:features}.ii. These two outputs were then concatenated to form the full state-space representation for the agent.



\subsection{Curriculum Learning}

Curriculum learning is a method for training machine learning models by gradually increasing the complexity of the task to be solved as well as the data samples used during training in order to improve training efficiency \cite{Soviany2021}.
In the reinforcement learning, curriculum learning is a vital component that ensures the agents receives positive rewards even during early stages of training when the policy is not fully developed.
This technique has enabled reinforcement learning agents to learn how to control a quadrupedal robot to walk on challenging terrain \cite{lee2020learning,rudin2021learningtowalk} and complete hiking trails \cite{miki2022wyldanimal} (policies were trained first on flat terrain, then progressively moved to more challenging ones with slopes and obstacles), control realistic bipedal robots in simulation to walk over stepping stones \cite{xie2020allsteps} (curriculum is used to generate courses of different complexities), and a 2D simulated obstacle course with different levels \cite{wang2019paired}.
Self-play or League-based training, is another form of curriculum learning whereby the agent learns from playing against previously trained versions of itself.  Agents trained under this strategy have had notable success in competitive environments
This strategy was previously used to train an agent to play the full-game of \textit{StarCraft II} and achieve grandmaster level, which translates to ranking above 99.8\% of the officially ranked human players.  \cite{vinyals2019grandmaster}.

\subsubsection{Automatic Curriculum Generation}

\begin{figure}
    \centering
    \subfloat[\label{fig:autocurriculum_steps_a}]{\includegraphics[width=0.24\textwidth]{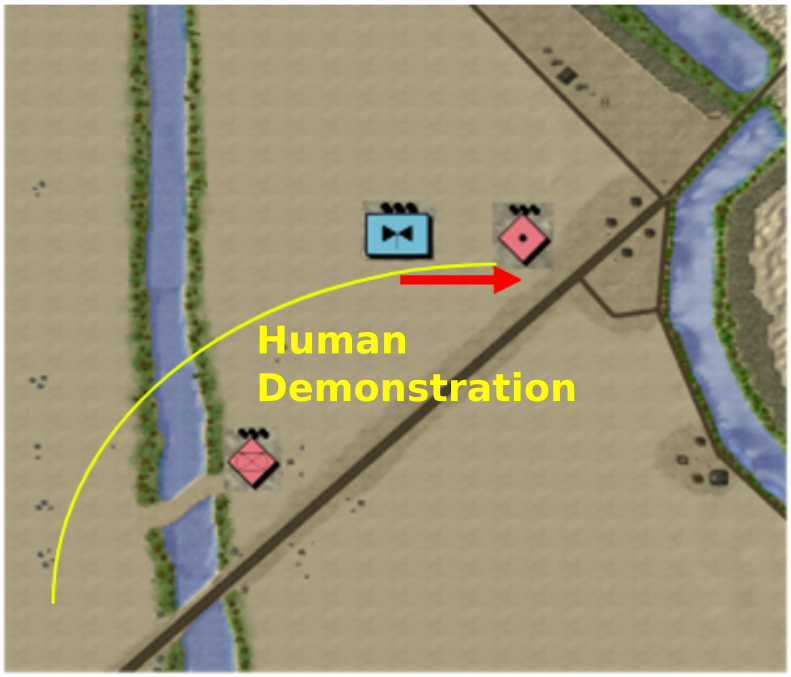}}\hfill
    \subfloat[\label{fig:autocurriculum_steps_b}]{\includegraphics[width=0.24\textwidth]{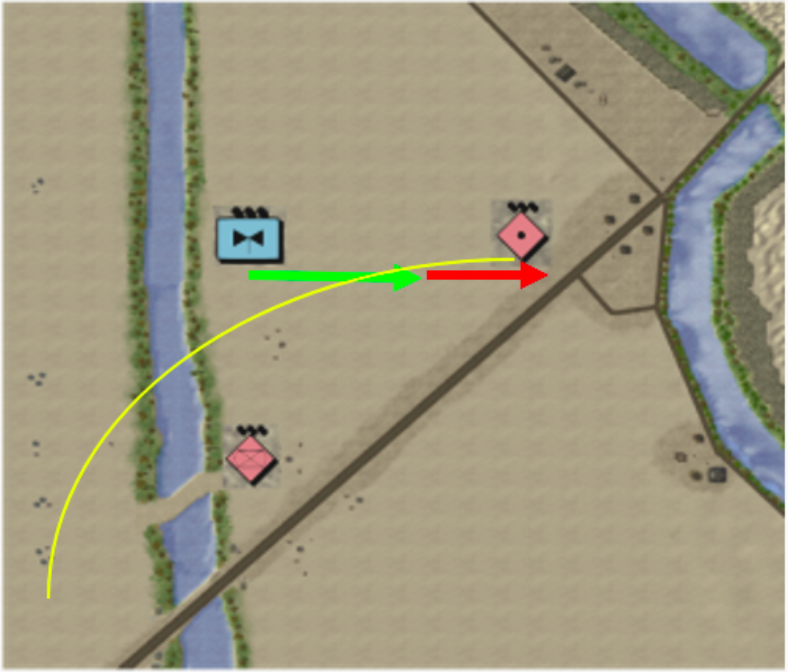}}\hfill
    \subfloat[\label{fig:autocurriculum_steps_c}]{\includegraphics[width=0.24\textwidth]{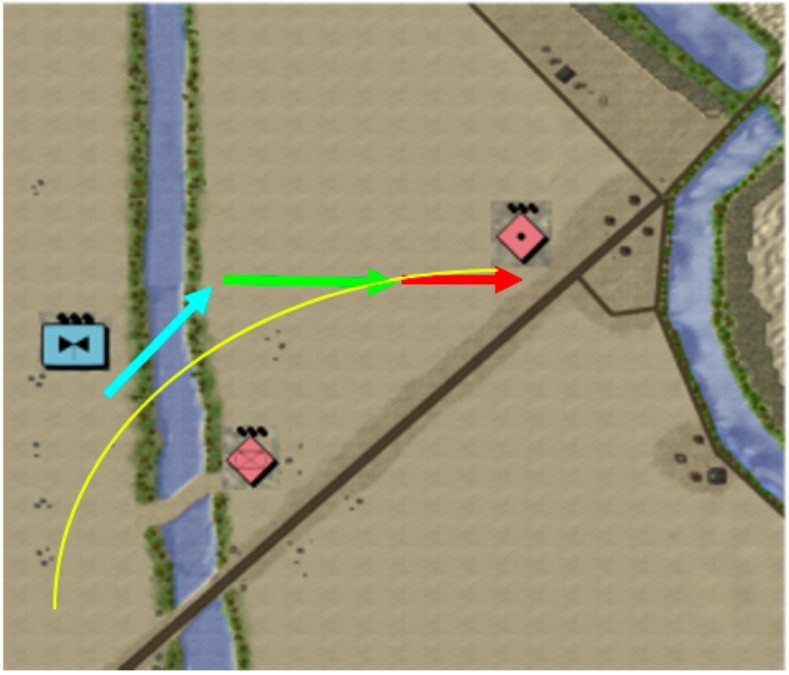}}\hfill
    \subfloat[\label{fig:autocurriculum_steps_d}]{\includegraphics[width=0.24\textwidth]{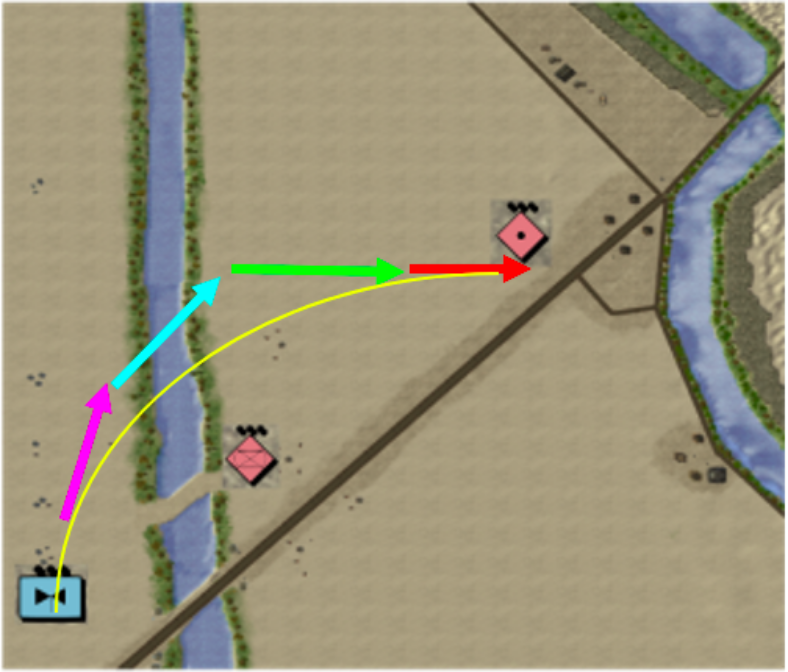}}
    \caption{Diagram illustrating how a single human demonstration is used to generate a curriculum to the reinforcement learning agent. The goal of the agent (Blue Force) is to avoid the Red Forces near the bridge and engage them at the corner of the map. the The agent starts learning from the end of the demonstrated trajectory \ref{fig:autocurriculum_steps_a}, represented by a yellow line. Once this final segment is mastered, as judge by the reward value, the agent starts to learn further from the end of the trajectory, as seen in \ref{fig:autocurriculum_steps_b} and \ref{fig:autocurriculum_steps_c}. This process repeats until the agent masters all segments and starts at the beginning of the demonstrated trajectory, as shown in \ref{fig:autocurriculum_steps_d}.}
    \label{fig:autocurriculum_steps}
\end{figure}

In automatic curriculum generation (ACL), the difficulty of the task is automatically controlled based on the current skill level of the agent during the training procedure. Self-play can be thought as a form of ACL since the agent learns continuously playing against itself. This results in a self-sustaining loop where the agent gets better as its opponent (i.e the agent itself) also gets better. However, for our task, we are unable to utilize traditional self-play since the Blue Force and Red Forces are heterogeneous and are composed of different StarCraft units.  
Instead, our main approach revolves around training an RL model using curriculum learning defined by a single human demonstration. We do this by starting the agent at the near end of the human demonstration, and then progressively making the task more difficult by rolling back through the human demonstration as the agent gets better and better, illustrated in Fig.~\ref{fig:autocurriculum_steps}.

In detail, this procedure is accomplished by the following scheme:
1) We record the sequence of states, actions and rewards (i.e., the trajectory) observed during the demonstration of the human user performing the given task.
2) Given this trajectory, the automatic curriculum generation starts by letting the agent complete only the last few steps of the task, as demonstrated by the human. 
This is done by first sampling a state from the end of the human demonstration from which the learning agent will execute its current policy. We record the trajectory resulting from this rollout, and train the policy by comparing the rewards received by the agent in this trajectory to the ones observed during demonstration.
We repeat this step until the agent is proficient at this part of the curriculum.
We consider this an easier task to be solved because the agent only needs to perform a few good actions to complete the task.
3) Once the agent has achieved comparable performance to the human, at the current stage of the curriculum, we then increase the difficulty of the task by setting the initial states of the agent further away from the end of the task along the trajectory generated generated by the human.
4) The agent continues following this scheme until it becomes proficient in solving the task from all states demonstrated by the human, completing the curriculum.

For our ACL agent, the exact point in the human demonstrated trajectory where we start the curriculum is represented by a Gaussian distribution with mean that ranges from 0 to 1 (0 to 100\% of the trajectory) and standard deviation of $1/6$.
During each rollout of the task, the starting point of the agent is sampled from this distribution and clipped between 0 and 100\%.
This helps the agents to experience diverse starting locations instead of overfitting to a single point, plus, ACL acts as a way to guide the exploration for an RL agent for efficient learning.

\section{EXPERIMENTS and RESULTS}

\subsection{Experimental Conditions} \label{sec:exp_cond}

In order to understand the benefits of ACL, we conducted the following four experiments: (i) \emph{Automatic Curriculum Learning with Image Representation}, (ii) \emph{Automatic Curriculum Learning with Vector Representation}, (iii) \emph{No Automatic Curriculum Learning (Traditional RL) with Image Representation}, and (iv) \emph{No Automatic Curriculum Learning (Traditional RL) with Vector Representation}.
The first and third experiments follow the network architecture shown in Fig.~\ref{fig:baseline}, while the second and forth experiments remove the Image Features from the input. Thus, we aim to understand if (a) ACL achieves more reward than traditional reinforcement learning and if (b) increasing the observation state space to include terrain information in image format improves learning. 

For each experimental condition, we trained our agent on the DoD High Performance Computing system with 35 parallel actor-learners for over $70$ million timesteps (around $112$ thousand simulated battles) against a built-in \emph{StarCraft II} bot operating on hand-crafted rules.
The traditional RL algorithms resets each episode so the agent starts at the beginning of the game, while the ACL experiments sample the start of the episode from a Gaussian distribution initially centered at 95\% of the human-demonstrated trajectory, that is, the start of each episode the battle is played out according to the first sampled percentage of the actions of the human demonstration, then the ACL agent takes over.
Once the ACL agent finishes at least 50 battles with a score similar to the human demonstration, the mean of the Gaussian distribution is rolled back through the curriculum in 20\% increments to provide progressively more and more difficult tasks of the ACL agent.

\subsection{Results}

\begin{figure}
    \centering
    \subfloat[\label{fig:reward_img_rep}]{\includegraphics[width=0.48\textwidth]{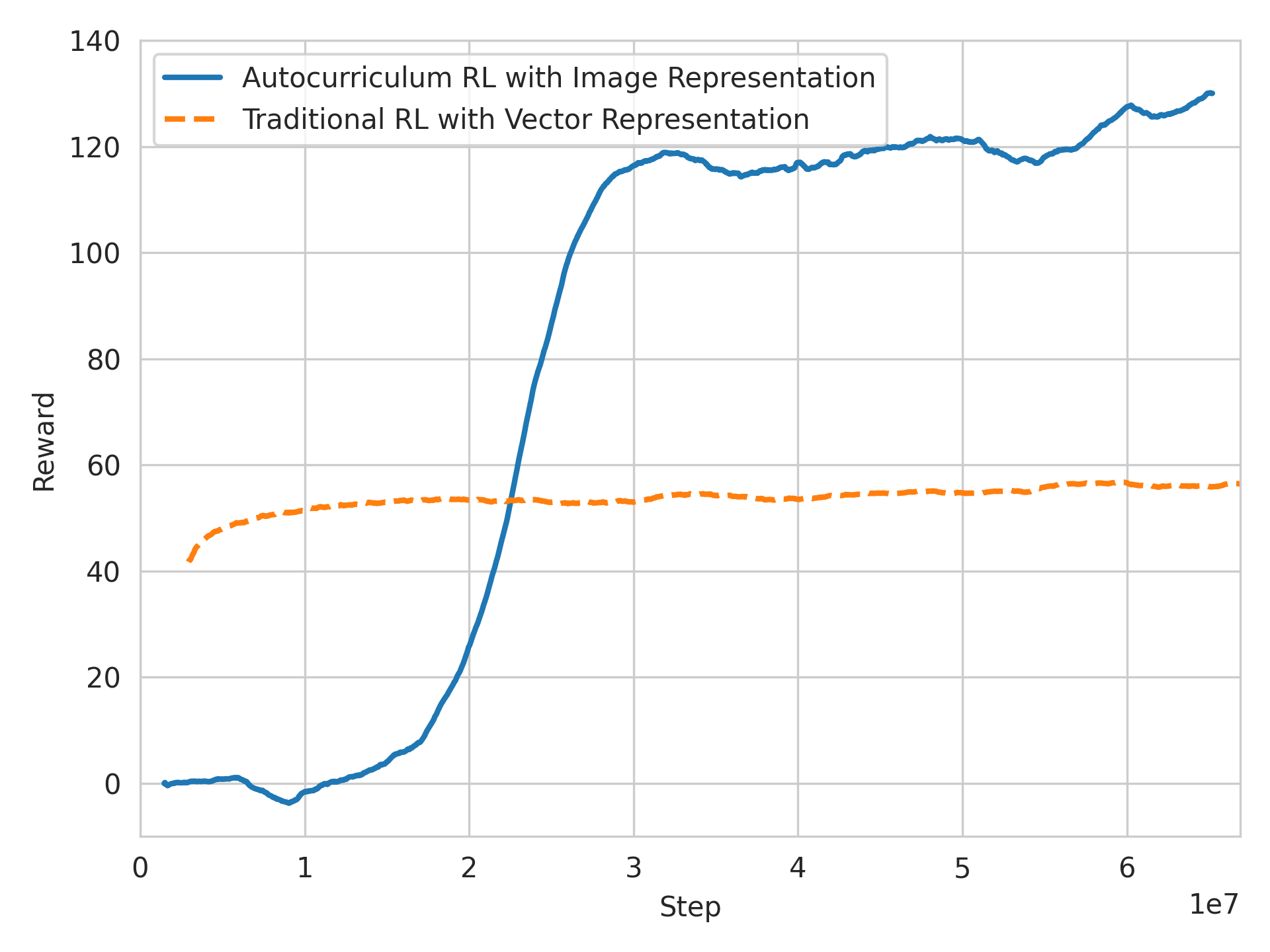}}\hfill
    \subfloat[\label{fig:reward_vec_rep}]{\includegraphics[width=0.48\textwidth]{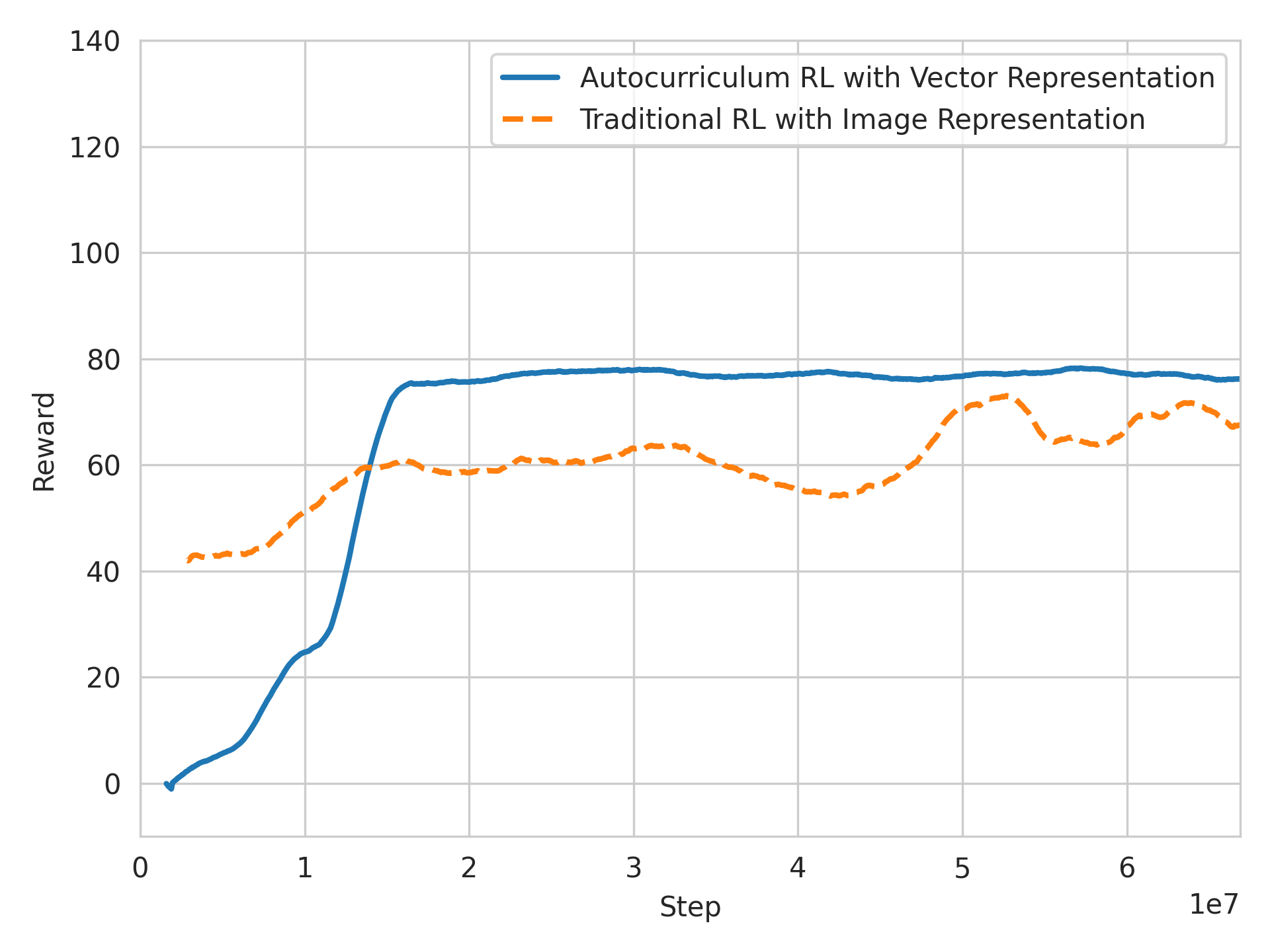}}
    \caption{Average episode reward received by the agent during training when using the curriculum automatically generated from human demonstrations (Autocurriculum RL) and when using traditional RL methods with no curriculum (Traditional RL) with the state information of the environment being represented as images (Fig.~\ref{fig:reward_img_rep}) or vectors (Fig.~\ref{fig:reward_vec_rep}).}
    \label{fig:reward_plot}
\end{figure}




Our first result compares agents that were trained using images (terrain information) within their observed states.
Fig.~\ref{fig:reward_img_rep} presents the average reward achieved by the agent for experiments (i) and (iii), \emph{Automatic Curriculum Learning with Image Representation} and \emph{No Automatic Curriculum Learning (Traditional RL) with Image Representation}, respectively.
As seen, initially, the reward achieved using ACL is much less than without ACL due to the agent starting at the end of the curriculum, as defined by the human-demonstrated trajectory.
In this case, the agent only needs to eliminate the last couple units of the opposing force, which does not result in a large reward. However, after $3$ million episodes, the ACL agent has moved further along the curriculum and has learned to eliminate the opposing force, resulting in a larger reward than without ACL.
Furthermore, the reward for the agent without ACL (Traditional RL) slowly increases, indicating that learning a complex task from the beginning, that is, searching over the full state space of the task, results in a agent that learns slower than an agent that is guided through a curriculum.

\begin{table}
\caption{Average casualties for each force.}
\centering
\begin{tabular}{c|c|cc|cc}
                & \multicolumn{1}{c|}{Human}      & \multicolumn{2}{c|}{Autocurriculum RL}          & \multicolumn{2}{c}{Traditional RL}        \\
                & \multicolumn{1}{c|}{Demonstration}      & \multicolumn{1}{c|}{Image Rep.} & Vector Rep. & \multicolumn{1}{c|}{Image Rep.} & Vector Rep. \\ \hline
Blue Force Casualties & \multicolumn{1}{c|}{$2.80$} & \multicolumn{1}{c|}{\textbf{$2.94$}}     & $10.66$          & \multicolumn{1}{c|}{$13.57$}         & $11.22$          \\
Red Force Casualties & \multicolumn{1}{c|}{$17.00$} & \multicolumn{1}{c|}{$13.17$}    & $13.08$          & \multicolumn{1}{c|}{$15.02$}         & $12.33$         
\end{tabular} \label{tab:casualties}
\end{table}

We also evaluated the average amount of casualties for the Blue force and the Red force at the end of training.
As seen in Table~\ref{tab:casualties}, learning with ACL achieves \textit{significantly} fewer Blue force casualties and higher Red force casualties when compared to traditional RL.
This also explains why the ACL agent achieves a larger reward since the Blue force casualties contribute to a large negative reward. Additionally, the ACL with image representation almost achieves the casualty rates of the human demonstration, showing that the curriculum helped guide the agent to learn a similar policy. 

Next, we compared experiments (ii) and (iv), \emph{Automatic Curriculum Learning with Vector Representation} and \emph{No Automatic Curriculum Learning (Traditional RL) with Vector Representation}, respectively, where the agent uses only vector representations for the observed state. As seen in Fig.~\ref{fig:reward_vec_rep}, similar to the behavior observed in Fig.~\ref{fig:reward_img_rep}, the reward for the ACL surpasses the reward without ACL after a few million training steps.
However, the gap between the two experiments is not as significant as the agents trained with image representations.
In the case of experiments where the agents are trained with only the vector representation of the environment, both agents result in similar Blue force and Red force casualties, with ACL slightly outperforming traditional RL. 

When we compare experiments (i) and (iii) against (ii) and (iv), we note that agents with image representations achieve a larger reward than agents with vector representations due to the resulting fewer Blue Force casualties and similar number for Red Force casualties.
This indicates that including additional information, i.e., terrain information, allows for the agent to learn a better policy. However, this result is at the cost of additional training time due to the complexity in the observational state space.

To further understand how the ACL policies were outperforming traditional RL and the differences from the human demonstration, we collected data on how the agents were commanding each coalition, as represented by their distance travelled (Table \ref{tab:distance}) and health percentage remaining at the end of the battle (Table \ref{tab:health}).
Looking at the policies that utilize terrain information (image representation as input), ACL policies maneuvered their units more often than traditional RL and the human demonstration.
For example, \emph{aviation} units (which are faster and able to attack both ground and air units, but with lower total health capacity) travelled more than five times more in ACL than traditional RL and three times more than the human demonstration.
All other coalitions travelled at least twice more in ACL. 
This could have led to higher fuel usage but units were able to eliminate more Red Forces and remain alive.
To illustrate this point, Table \ref{tab:health} shows that mortar units commanded by ACL received no damage.
All coalitions commanded by ACL, on average, finished the battles with more health percentage remaining when compared to all other conditions.

\begin{table}
\caption{Average distance travelled for each coalition of the Blue force.}
\centering
\begin{tabular}{c|c|cc|cc}
                   & \multicolumn{1}{c|}{Human}   & \multicolumn{2}{c|}{Autocurriculum RL}          & \multicolumn{2}{c}{Traditional RL}        \\
                   & \multicolumn{1}{c|}{Demonstration}   & \multicolumn{1}{c|}{Image Rep.} & Vector Rep. & \multicolumn{1}{c|}{Image Rep.} & Vector Rep. \\ \hline
Aviation & \multicolumn{1}{c|}{$295.91$} & \multicolumn{1}{c|}{\textbf{$1041.99$}}     & $243.88$          & \multicolumn{1}{c|}{$178.30$}         & $351.90$          \\
Mechanized Infantry & \multicolumn{1}{c|}{$350.00$} & \multicolumn{1}{c|}{$818.90$}    & $495.45$          & \multicolumn{1}{c|}{$443.14$}         & $287.49$         \\
Mortars & \multicolumn{1}{c|}{$88.19$}  & \multicolumn{1}{c|}{$218.02$}    & $209.05$          & \multicolumn{1}{c|}{$123.85$}         & $92.09$            \\
Scouts & \multicolumn{1}{c|}{$83.73$}  & \multicolumn{1}{c|}{$235.60$}    & $83.72$          & \multicolumn{1}{c|}{$36.02$}         & $46.90$               \\
Tanks & \multicolumn{1}{c|}{$597.10$}  & \multicolumn{1}{c|}{$2281.20$}    & $1330.18$          & \multicolumn{1}{c|}{$1381.22$}         & $1763.06$ 
\end{tabular}
 \label{tab:distance}
\end{table}

\begin{table}
\caption{Average health percentage remaining for each coalition of the Blue force.}
\centering
\begin{tabular}{c|c|cc|cc}
                    & \multicolumn{1}{c|}{Human}  & \multicolumn{2}{c|}{Autocurriculum RL}          & \multicolumn{2}{c}{Traditional RL}        \\
                    & \multicolumn{1}{c|}{Demonstration}  & \multicolumn{1}{c|}{Image Rep.} & Vector Rep. & \multicolumn{1}{c|}{Image Rep.} & Vector Rep. \\ \hline
Aviation & \multicolumn{1}{c|}{$59.78$} & \multicolumn{1}{c|}{\textbf{$29.48$}}     & $1.15$          & \multicolumn{1}{c|}{$0.63$}         & $17.65$          \\
Mechanized Infantry & \multicolumn{1}{c|}{$48.16$} & \multicolumn{1}{c|}{$59.58$}    & $12.42$          & \multicolumn{1}{c|}{$6.32$}         & $8.87$         \\
Mortars & \multicolumn{1}{c|}{$96.62$} & \multicolumn{1}{c|}{$100.00$}    & $11.60$          & \multicolumn{1}{c|}{$1.90$}         & $15.01$            \\
Scouts & \multicolumn{1}{c|}{$22.30$} & \multicolumn{1}{c|}{$29.00$}    & $0.12$          & \multicolumn{1}{c|}{$0.00$}         & $2.96$               \\
Tanks & \multicolumn{1}{c|}{$28.77$} & \multicolumn{1}{c|}{$35.57$}    & $14.03$          & \multicolumn{1}{c|}{$4.96$}         & $15.97$ 
\end{tabular} \label{tab:health}
\end{table}

\section{DISCUSSION}

Curriculum Learning is a technique designed to make learning complex problems easier by designing a curriculum of simple tasks and concepts to improve the speed and efficiency of learning
Curriculum learning has been utilized in the reinforcement learning field in order to improve the sample efficiency of RL training, however, manually designing a curriculum is often challenging. In this paper, we utilize human demonstrations to generate a curriculum automatically (i.e. automatic curriculum learning (ACL)) and then use that curriculum to speed up reinforcement learning on the complex environment of \textit{StarCraft II}. We show that with our curriculum learning agent, we are able to significantly improve training speed and performance compared to traditional RL agents in \textit{StarCraft II}. 


In reinforcement learning, the state-space that is made available to the agent is often a critical factor in determining the complexity of the learning problem. In this paper, we analyzed two different state-space representations for our learning agents, a high dimension, image-based representation and a low-dimension, vector-based representation. The image representation consists of game-screen images of the \textit{StarCraft II} environment that contain unit information and terrain information. The vector representation is a much more compact representation that contains the entire state of the game in a vectorized format (i.e. unit position, health, etc.), however, it does not contain any terrain information. Generally, there is a trade-off between the higher-complexity image representation and lower-dimensional vector representation in terms of both training time and overall policy performance. Based on our results, as seen in Figs.~\ref{fig:reward_img_rep} and \ref{fig:reward_vec_rep}, we show that despite the larger state-space, both traditional RL and ACL agents achieve a higher overall reward using image representation when compared to the vector representation-based agents. Interestingly, as shown in Table~\ref{tab:casualties}, we see that for the Autocurriculum RL agent trained with images, the agent was able to dramatically reduce the number of Blue Force casualties to only 3, compared to the 10-13 casualties that the vectorized representation produced. The results in Table~\ref{tab:distance} show that the image based ACL agent utilizes the aviation units more often, as indicated by the amount of distance traveled. These are powerful units since they are fast and can attack enemy air and ground forces. Although more analysis needs to be performed, we speculate that having access to terrain information allows for the ACL agent to better follow the human guided curriculum, which relies on heavy use of the aviation units, which ultimately leads to fewer Blue Force casualties. 

\subsection{Limitations and Future Work}

Although we have shown that there is a benefit to performance in terms of overall reward and performance achieved by the ACL agent compared to traditional RL, there are several limitations of the current work that we highlight here as the topic of future studies. Firstly, as with almost all deep reinforcement learning work, training RL agents on these tasks is incredibly difficult and requires extensive hyper-parameter turning to achieve proper convergence. We found that our approach of using curriculum learning to train RL agents is not immune to this problem. 

With curriculum learning, one of the difficulties is deciding when and how the RL agent should progress through the curriculum. One limitation in the current work is that we utilized a rather naive and straightforward curriculum step policy of just setting a fixed performance threshold before allowing the agent to progress to the next step in the curriculum. This leads to problems of the agent potentially getting stuck at a certain part of the curriculum and never progressing because the threshold is too high. This ultimately slows down learning to an extent where curriculum learning is no longer useful or practical. Future work could involve developing more intelligent strategies for traversing the curriculum. 

\acknowledgments 
 
This work was supported in part by high-performance computer time and resources from the DoD High Performance Computing Modernization Program.
This work was also sponsored by the Army Research Laboratory and was accomplished partly under Cooperative Agreement Number W911NF-20-2-0114. The views and conclusions contained in this document are those of the authors and should not be interpreted as representing the official policies, either expressed or implied, of the Army Research Laboratory or the U.S. Government. The U.S. Government is authorized to reproduce and distribute reprints for Government purposes notwithstanding any copyright notation herein.

\bibliography{report} 
\bibliographystyle{spiebib} 

\end{document}